\documentclass[pdflatex,sn-mathphys-num]{sn-jnl}

\usepackage{graphicx}
\usepackage{multirow}
\usepackage{amsmath,amssymb,amsfonts}
\usepackage{amsthm}
\usepackage{mathrsfs}
\usepackage[title]{appendix}
\usepackage{xcolor}
\usepackage{textcomp}
\usepackage{manyfoot}
\usepackage{booktabs}
\usepackage{algorithm}
\usepackage{algorithmicx}
\usepackage{algpseudocode}
\usepackage{listings}

\begin{document}

\title[PatchCTG: Patch Cardiotocography Transformer for Antepartum Fetal Health Monitoring]{PatchCTG: Patch Cardiotocography Transformer for Antepartum Fetal Health Monitoring}

\author*{\fnm{M. Jaleed} \sur{Khan}}\email{jaleed.khan@wrh.ox.ac.uk}
\author{\fnm{Manu} \sur{Vatish}}\email{manu.vatish@wrh.ox.ac.uk}
\author*{\fnm{Gabriel} \sur{Davis Jones}}\email{gabriel.jones@wrh.ox.ac.uk}
\affil{\orgdiv{Nuffield Department of Women's \& Reproductive Health (NDWRH)}, \orgname{University of Oxford}, \orgaddress{\street{Women's Centre, John Radcliffe Hospital}, \city{Oxford}, \postcode{OX3 9DU}, \country{United Kingdom}}}

\abstract{Antepartum Cardiotocography (CTG) is vital for fetal health monitoring, but traditional methods like the Dawes-Redman system are often limited by high inter-observer variability, leading to inconsistent interpretations and potential misdiagnoses. This paper introduces PatchCTG, a transformer-based model specifically designed for CTG analysis, employing patch-based tokenisation, instance normalisation and channel-independent processing to capture essential local and global temporal dependencies within CTG signals. PatchCTG was evaluated on the Oxford Maternity (OXMAT) dataset, comprising over 20,000 CTG traces across diverse clinical outcomes after applying the inclusion and exclusion criteria. With extensive hyperparameter optimisation, PatchCTG achieved an AUC of 77\%, with specificity of 88\% and sensitivity of 57\% at Youden's index threshold, demonstrating adaptability to various clinical needs. Testing across varying temporal thresholds showed robust predictive performance, particularly with finetuning on data closer to delivery, achieving a sensitivity of 52\% and specificity of 88\% for near-delivery cases. These findings suggest the potential of PatchCTG to enhance clinical decision-making in antepartum care by providing a reliable, objective tool for fetal health assessment. The source code is available at \href{https://github.com/jaleedkhan/PatchCTG}{https://github.com/jaleedkhan/PatchCTG}}


\maketitle

\section{Introduction}\label{sec1}

Antepartum Cardiotocography (CTG) plays a pivotal role in fetal health monitoring, serving as a critical assessment tool in prenatal care. Using ultrasound-based techniques to record Fetal Heart Rate (FHR) and uterine activity, CTG provides clinicians with data on fetal well-being through the examination of heart rate variability and response patterns to uterine contractions. Established methods like the Dawes-Redman (DR) computerised CTG system \cite{pardey2002computer} offer valuable criteria for interpreting CTG patterns, enhancing clinical decisions that help mitigate risks of adverse outcomes such as neonatal acidaemia, hypoxia and stillbirth \cite{ayres2015figo}. Despite its widespread adoption in clinical settings, CTG analysis suffers from high intra- and inter-observer variability. Studies indicate that clinical assessments can overlook 35-92\% of FHR patterns \cite{Gagnon1993,Todros1996}, and inter-observer agreement may be as low as 29\%, with false positive rates reaching up to 60\% \cite{Beaulieu1982,Borgatta1988}. These limitations underscore the necessity for more reliable, objective methods of fetal monitoring.

Recent advancements in artificial intelligence and machine learning, particularly deep learning, have demonstrated considerable potential in automating and improving the accuracy of CTG interpretation. By leveraging deep learning models, researchers have advanced the detection of adverse outcomes in CTG signals through feature extraction, noise reduction and classification tasks, providing more consistent assessments than manual interpretation \cite{aeberhard2023artificial}. Transformers excel at handling sequential data, including biomedical time series, due to their ability to capture complex temporal dependencies by dynamically learning correlations across input elements, which makes them highly promising for CTG analysis \cite{barnova2024artificial}. With self-attention mechanisms, Transformers capture complex temporal dependencies in CTG data, focusing on relevant segments of FHR and uterine activity patterns. 

Despite these advancements, several challenges persist in applying deep learning to CTG-based fetal health monitoring. Existing models fall short in capturing the physiological responses in CTG, largely due to signal variability across patients, monitoring conditions and clinical contexts \cite{aeberhard2024introducing}. Issues like insufficient data diversity, high computational costs and a lack of generalizability across different clinical settings often limit the performance of deep learning models on CTG data. Addressing these challenges requires specialised models that can adapt to different temporal patterns while maintaining robust performance. 

In this paper, we introduce Patch Cardiotocography Transformer (PatchCTG), a patch-based transformer model designed to classify adverse and normal outcomes in antepartum CTG recordings reliably. The PatchCTG model builds on recent advancements in patch-based Transformers for time series \cite{nie2022forecasting}, which demonstrate promising performance in sequence compression and feature representation by segmenting signals into patches. Unlike traditional CTG analysis approaches, PatchCTG applies instance normalisation and channel-independent processing to manage distribution shifts and capture the distinct temporal dynamics of FHR and uterine activity. By leveraging patch-based tokenisation and self-attention, PatchCTG provides enhanced computational efficiency and adaptability to longer temporal windows, making it well-suited for CTG data, where signal length and variability pose significant modelling challenges. The main contributions of this study are as follows:

\begin{itemize}
    \item We introduce PatchCTG, a Transformer-based architecture tailored for CTG signals through patch-based segmentation, instance normalisation and channel-independent processing. This design effectively captures both local and global temporal dependencies by segmenting signals into patches, mitigating distribution shifts through instance normalisation and allowing separate modelling of FHR and uterine contraction channels. These architectural choices address the non-stationarity of CTG data, enabling more accurate classification of adverse and normal outcomes.
    \item The PatchCTG model is rigorously evaluated on a subset of the extensive Oxford Maternity (OXMAT) dataset \cite{khan2024oxmat}, which includes over 20,000 CTG traces with diverse clinical outcomes. Our experimentation involves cohort balancing, sequence standardisation and multiple temporal thresholds to ensure model robustness and clinical applicability. Detailed results demonstrate the promising performance of PatchCTG across various classification thresholds and time windows relative to delivery.
    \item We employed the Optuna hyperparameter optimisation framework \cite{akiba2019optuna} to identify the best configuration of PatchCTG for fetal health classification. This systematic approach finetunes model depth, attention heads, embedding dimensions, dropout rates and other key parameters, ensuring that PatchCTG achieves high predictive performance while maintaining generalizability across CTG samples and cohort variations.
    \item We benchmarked the performance of PatchCTG against the traditional Dawes-Redman algorithm \cite{pardey2002computer} and an optimised hybrid deep learning model, providing a comprehensive comparison that underscores the advantages of PatchCTG in handling temporal dependencies and delivering clinically relevant sensitivity and specificity.
\end{itemize}

The remainder of the paper is structured as follows: Section \ref{sec2} reviews related work in antepartum fetal monitoring using machine learning. Section \ref{sec3} presents the proposed PatchCTG model architecture, and Section \ref{sec4} presents the experimental setup and results, including dataset preprocessing, hyperparameter optimisation, performance evaluation and benchmark comparison. Section \ref{sec5} discusses the experimental results of PatchCTG, its potential impact on clinical practice and future directions for its development, which is followed by the conclusion in Section \ref{sec6}.

\section{Related Work}\label{sec2}

The Dawes-Redman (DR) system has long served as the gold standard in electronic fetal monitoring, providing a rule-based algorithm for interpreting CTG signals by analysing FHR variability and responses to uterine contractions \cite{pardey2002computer}. Despite its wide adoption, studies reveal that CTG interpretation using DR criteria is often marred by substantial inter-observer variability, with agreement rates between clinicians ranging from 35\% to 92\%, largely depending on experience levels \cite{Gagnon1993, Todros1996}. This subjectivity raises reliability concerns, as inconsistent interpretations can lead to misclassification and potentially adverse outcomes. Neppelenbroek et al. \cite{neppelenbroek2024inter} further underscored this issue, showing that only professionals with high training consistency in controlled environments achieved satisfactory inter- and intra-observer agreement, which varied significantly from 64\% to 98\%. Jones et al. \cite{jones2024performance} evaluated the performance of the DR algorithm in predicting adverse outcomes, using 4,196 antepartum FHR recordings and excluding those with incomplete data or terminated analyses. Their findings indicated that while the DR algorithm showed high sensitivity (91.7\%) for detecting fetal well-being, its specificity for adverse outcomes was low (15.6\%), limiting its predictive utility in high-risk pregnancies.

To mitigate the subjectivity and variability of manual CTG assessment, machine learning approaches have been proposed as alternatives to improve reliability and interpretative accuracy. Traditional ML methods such as the EMD-SVM model by Krupa et al. \cite{krupa2011antepartum} employed Empirical Mode Decomposition (EMD) for feature extraction and Support Vector Machine (SVM) classification, achieving 87\% accuracy with a high agreement (kappa value 0.923) with expert evaluations. Georgieva et al. \cite{georgieva2013artificial} employed an ensemble of Artificial Neural Networks (ANNs) for adverse outcome prediction, achieving a sensitivity of 60.3\% and specificity of 67.5\%. Fei et al. \cite{fei2020automatic} developed an Adaptive Neuro-Fuzzy Inference System (FCM-ANFIS) that achieved 96.39\% accuracy, outperforming conventional classifiers. Chen et al. \cite{chen2021intelligent} introduced a Deep Forest model that handled overlapping normal and suspicious classifications with 92.64\% accuracy on the UCI dataset \cite{uci2024ctg}. However, while these traditional approaches demonstrated some promise, their limited feature extraction capabilities often restricted their generalisation across diverse datasets.

With advancements in deep learning, more sophisticated models like Convolutional Neural Networks (CNNs) and Long Short-Term Memory (LSTM) networks gained traction in CTG analysis, allowing for improved temporal feature extraction and pattern recognition in FHR variability. Petrozziello et al. \cite{petrozziello2019multimodal} applied Multimodal CNNs to predict fetal distress by processing FHR and uterine contraction data, achieving a True Positive Rate (TPR) of 53\% at a 15\% False Positive Rate (FPR). Ogasawara et al. \cite{ogasawara2021deep} proposed CTG-net, a three-layer CNN model that achieved an AUC of 0.73±0.04, outperforming SVM and k-means clustering. Xiao et al. \cite{xiao2022deep} used a multiscale CNN-BiLSTM model to capture spatial and temporal features, reaching a sensitivity of 61.97\% and specificity of 73.82\% on the CTU-UHB dataset \cite{chudacek2014open}. Fei et al. \cite{fei2022intelligent} developed a Multimodal Bidirectional Gated Recurrent Unit (MBiGRU) network, achieving an AUC of 93.27\%. Although these DL approaches showed an improved ability to extract intricate temporal patterns, their capacity to capture long-range dependencies in non-stationary CTG data and to generalise was still limited.

In recent years, hybrid models have emerged, integrating diverse neural architectures to harness complementary strengths. For instance, Spairani et al. \cite{spairani2022deep} combined a Multi-Layer Perceptron (MLP) and CNN for mixed quantitative and image-derived inputs, achieving an accuracy of 80.1\% but showing limited sensitivity. Feng et al. \cite{feng2023hybrid} applied an ensemble of SVM, eXtreme Gradient Boosting (XGB) and random forest models, reaching 0.9539 accuracy on the UCI dataset. Zhang et al. \cite{zhang2023dt} developed DT-CTNet, which combined an XGBoost ensemble and CNN-based tracking, achieving a diagnostic accuracy of 96.3\%. Chen et al. \cite{chen2024dannmctg} introduced an Unsupervised Domain Adaptation (UDA) model, DANNMCTG, to handle cross-device discrepancies, achieving an accuracy of 71.25\%. These models have demonstrated potential for enhanced generalisation and interpretability; however, their performance frequently depends on extensive feature engineering and high computational demands.

The emergence of self-attention mechanisms and transformer-based models has shifted the landscape in time-series classification, offering enhanced capacity to capture complex temporal dependencies. For CTG data, self-attention models have shown the potential to address the limitations of CNNs and RNNs by dynamically focusing on relevant parts of the signal. Asfaw et al. \cite{asfaw2024gated} introduced a Gated Convolutional Multi-Head Attention (GCMHA) model, which combined CNNs with attention mechanisms to refine temporal dependencies, achieving a sensitivity of 49.08\% at a 15\% FPR. Wu et al. \cite{wu2024etcnn} proposed the Ensemble Transformer-Convolutional Neural Network (ETCNN), designed to capture both short and long-term features by segmenting FHR patterns into acceleration and deceleration phases. ETCNN demonstrated improved segmentation accuracy, achieving an 80.68\% accuracy for accelerations and a 78.24\% accuracy for decelerations.

Beyond general transformer-based models, emerging research \cite{nie2022forecasting} suggests that patch-based transformers offer unique advantages for time-series analysis. By segmenting sequences into patches, these models reduce input dimensionality while preserving local and global temporal information, improving feature extraction and computational efficiency. Patch-based segmentation has enhanced computational efficiency and improved feature representation in complex, non-stationary data, making it a promising approach for CTG classification. The design of patch-based transformers aligns well with CTG data requirements, where signal length and variability are considerable challenges for conventional approaches.

\section{Proposed Method}\label{sec3}

The Patch Cardiotocography Transformer (PatchCTG) is a transformer-based architecture designed to classify CTG signals into binary outcomes: adverse or normal. It builds upon the time series forecasting transformer architecture~\cite{nie2022patchtst}, which has been adapted for the specific task of CTG classification. PatchCTG focuses on time series classification using FHR and TOCO signals, each consisting of $L = 960$ time steps (corresponding to one hour of recording). PatchCTG efficiently extracts temporal dependencies from CTG signals through a workflow that includes instance normalisation for mitigating distribution shifts, patching for sequence compression, channel-independent processing, a Transformer backbone for temporal modelling and a classification head with pooling, dense layer and sigmoid activation for prediction.

In time-series analysis, particularly in clinical datasets, the characteristics of input signals can vary significantly due to various factors such as variations between patients and recording conditions. To mitigate the resulting distribution shift effects between training and testing data, PatchCTG adopts \textbf{instance normalisation}, which has proven effective in reducing such distribution issues ~\cite{ulyanov2016instance,kim2022reversible}. Instance normalisation independently standardises each univariate channel (FHR or TOCO) to have zero mean and unit variance. It recalculates mean and variance statistics for each sequence during inference, which helps to reduce patient-to-patient variability and ensures robustness across various recording conditions. Specifically, each univariate channel (FHR or TOCO) is independently standardised to have zero mean and unit variance. Given an input time series $\mathbf{x}^{(i)} = (x_1^{(i)}, x_2^{(i)}, \dots, x_L^{(i)})$, the normalised version $\tilde{\mathbf{x}}^{(i)}$ is computed as: 

\begin{equation}
    \tilde{\mathbf{x}}^{(i)} = \frac{\mathbf{x}^{(i)} - \mu^{(i)}}{\sigma^{(i)}},
\end{equation}

where $\mu^{(i)}$ and $\sigma^{(i)}$ are the mean and standard deviation of $\mathbf{x}^{(i)}$, respectively. This ensures that the model is robust to scale variations and more stable during training, which is essential for effective learning from clinical data, often involving different baselines for each patient and varied signal characteristics.

The PatchCTG method adopts a patching mechanism, which segments each univariate signal into a sequence of patches, inspired by the success of patch-based strategies in time-series forecasting~\cite{nie2022patchtst,zerveas2021transformer}. Patching effectively captures both local and global temporal trends, reduces input sequence length, and facilitates smoother temporal transitions in medical time series where gradual physiological changes occur. Given an input univariate signal $\mathbf{x}^{(i)}$, the patching mechanism divides $\mathbf{x}^{(i)}$ into non-overlapping or overlapping patches of fixed length $P$. Specifically, a patch $\mathbf{p}_j^{(i)}$ of the signal is defined as:

\begin{equation}
    \mathbf{p}_j^{(i)} = (x_{jS+1}^{(i)}, x_{jS+2}^{(i)}, \dots, x_{jS+P}^{(i)}), \quad j = 0, 1, \dots, N-1,
\end{equation}

where $S$ is the stride (step size), $P$ is the patch length, and $N$ is the total number of patches given by:

\begin{equation}
    N = \left\lfloor \frac{L - P}{S} \right\rfloor + 1.
\end{equation}

The stride $S$ controls the overlap between consecutive patches. By adjusting the stride, PatchCTG can create overlapping patches ($S < P$) to capture smoother transitions across temporal segments or non-overlapping patches ($S = P$) to focus on distinct episodes. Patching reduces the input sequence length by representing each patch as a single token, thereby enhancing computational efficiency. Each patch also retains local semantic information, which is critical for understanding physiological trends and events, such as identifying patterns in FHR that correlate with uterine contractions.

PatchCTG employs a \textbf{channel-independent Transformer encoder} for each univariate signal (FHR or TOCO). By processing each channel independently, the model learns unique temporal dynamics for each physiological signal before combining them for classification. The input encoding begins by projecting each patch from its original input space into a higher-dimensional latent space using a linear transformation:

\begin{equation}
    \mathbf{z}_j^{(i)} = \mathbf{W}_P \mathbf{p}_j^{(i)}, \quad \mathbf{W}_P \in \mathbb{R}^{P \times d},
\end{equation}

where $d$ is the latent dimensionality of the patch representation. To preserve temporal information, positional encodings $\mathbf{W}_{\text{pos}} \in \mathbb{R}^{N \times d}$ are added to each patch representation, resulting in:

\begin{equation}
    \mathbf{e}_j^{(i)} = \mathbf{z}_j^{(i)} + \mathbf{W}_{\text{pos}}^{(j)},
\end{equation}

where $\mathbf{e}_j^{(i)}$ represents the encoded patch with temporal information. This positional encoding ensures that the model can learn to interpret the sequential changes in CTG signals, which is important for understanding FHR decelerations or accelerations in response to uterine contractions.

The \textbf{Transformer backbone} consists of $L$ encoder layers, each comprising two sub-layers:

\begin{itemize}
    \item Multi-Head Self-Attention (MHSA): The multi-head self-attention mechanism enables PatchCTG to learn relationships between different patches within a given signal, providing a comprehensive view of temporal dependencies across the time series~\cite{zerveas2021transformer}. Given query, key, and value matrices $\mathbf{Q}$, $\mathbf{K}$, and $\mathbf{V}$, the attention output $\mathbf{A}$ is computed as:

    \begin{equation}
        \text{Attention}(\mathbf{Q}, \mathbf{K}, \mathbf{V}) = \text{softmax}\left( \frac{\mathbf{Q} \mathbf{K}^T}{\sqrt{d_k}} \right) \mathbf{V},
    \end{equation}

    where $d_k$ is the dimensionality of the key vectors, and the scaling factor $\frac{1}{\sqrt{d_k}}$ ensures numerical stability.
    
    \item Feed-Forward Network (FFN): The feed-forward network is applied to each output of the MHSA. It consists of two linear transformations with a non-linearity in between (Gaussian Error Linear Unit, GELU). Given an input vector $\mathbf{h}_i$, the FFN output $\mathbf{f}_i$ is given by:

    \begin{equation}
        \mathbf{f}_i = \text{GELU}(\mathbf{W}_1 \mathbf{h}_i + \mathbf{b}_1) \mathbf{W}_2 + \mathbf{b}_2,
    \end{equation}

    where $\mathbf{W}_1$, $\mathbf{W}_2$ are learnable weight matrices, and $\mathbf{b}_1$, $\mathbf{b}_2$ are biases. Residual connections and layer normalization are employed to stabilize training and facilitate gradient flow across multiple layers.
\end{itemize}

PatchCTG handles missing values using a masking mechanism, which prevents the model from learning spurious relationships from incomplete data during attention computations. After processing the input patches through the Transformer backbone, PatchCTG applies \textbf{global average pooling} across the time dimension. Given the output representations $\{\mathbf{e}_j^{(i)}\}_{j=1}^N$, global average pooling computes:

\begin{equation}
    \mathbf{g}^{(i)} = \frac{1}{N} \sum_{j=1}^N \mathbf{e}_j^{(i)},
\end{equation}

where $\mathbf{g}^{(i)}$ is the aggregated feature representation for each input sequence. By using global pooling, PatchCTG extracts meaningful summary statistics across the entire time horizon of each channel, capturing short-term variations and long-term trends, both of which may have clinical relevance in determining adverse outcomes. The pooled representation is then passed to a \textbf{classification head}, consisting of a dense layer and a sigmoid activation function, which maps the aggregated features to a single output value:

\begin{equation}
    y^{(i)} = \sigma(\mathbf{W}_c \mathbf{g}^{(i)} + b_c),
\end{equation}

where $\mathbf{W}_c$ and $b_c$ are learnable parameters, and $\sigma(\cdot)$ is the sigmoid activation function. The output $y^{(i)}$ represents the probability that the input CTG corresponds to an adverse outcome, with a threshold of 0.5 used to assign a binary class label (adverse or normal). The use of the global average pooling layer ensures that the final classification is informed by the entire temporal trajectory of each CTG signal, rather than focusing only on specific patches. This design is particularly important in medical time-series analysis, where both long-term trends and short-term fluctuations can be indicative of clinical outcomes.

The training objective for PatchCTG is formulated as a \textbf{binary cross-entropy loss} function, which is well-suited for the binary classification problem. Given the predicted probability $\hat{y}$ and the true class label $y \in \{0, 1\}$, the binary cross-entropy loss $\mathcal{L}$ is defined as:

\begin{equation}
    \mathcal{L} = -\left( y \log(\hat{y}) + (1 - y) \log(1 - \hat{y}) \right).
\end{equation}

The model parameters are optimised to minimise this loss over the training data. 

\section{Experiments and Results}\label{sec4}

\subsection{Data Preprocessing and Organisation}

The dataset used in this study was sourced from the Oxford Maternity (OXMAT) dataset~\cite{khan2024oxmat}, a comprehensive repository of CTG traces and maternal-neonatal health records collected from the Oxford University Hospitals maternity database at John Radcliffe Hospital. The OXMAT dataset contains over 211,000 CTGs, collected from more than 250,000 pregnancies between January 1991 and February 2024. Alongside CTG signals, the dataset includes over 250 clinical variables, which cover a range of maternal and neonatal outcomes such as Apgar scores, cord blood gas (CBG) values, birthweights, delivery types, medications and other related health parameters. For the development of PatchCTG, we adopted the dataset preprocessing methodology described in~\cite{davis2024performance} for cohort development and outcome categorisation. Raw digital CTG traces were extracted from singleton pregnancies between gestational weeks \(37^{+0}\) and \(41^{+6}\). The preprocessing involved removing CTG traces that were missing more than 30\% of their signal information or had aborted Dawes-Redman analysis before evaluation. Only traces that had undergone successful Dawes-Redman analysis were included.

To establish the Adverse Pregnancy Outcome (APO) cohort, traces acquired within 7 days prior to delivery were selected to ensure that CTG patterns used for classification were temporally related to the outcome. The adverse pregnancy outcomes considered included acidaemia, stillbirth, asphyxia, extended Special Care Baby Unit (SCBU) admission, Hypoxic-Ischaemic Encephalopathy (HIE), low Apgar score and neonatal resuscitation at delivery. These outcomes were chosen based on their clinical significance and correlation with neonatal health risks. To develop the Normal Pregnancy Outcome (NPO) cohort, inclusion and exclusion criteria were applied to identify traces with positive outcomes. Pregnancies in the NPO cohort included liveborn singleton babies with gestational age between \(37^{+0}\) and \(41^{+6}\) weeks, normal umbilical cord blood gas measurements, acceptable Apgar scores, and no major complications (e.g., emergency caesarean section or neonatal resuscitation). Traces from the NPO cohort were then matched to the APO cohort using one-to-one propensity score matching, controlling for key factors such as gestational age, maternal age, BMI, fetal sex, parity and monitoring time prior to delivery. 

After applying the inclusion and exclusion criteria and performing propensity score matching, we obtained a cohort consisting of 19,462 CTG traces (9,731 for the NPO group and 9,731 for the APO group). An 80-20 split was applied to this cohort for training and validation purposes. Additional preprocessing was performed to prepare the CTG signals for modelling. The FHR signal was adjusted to a range of $[50, 250]$ beats per minute, while the uterine contraction (TOCO) signal was adjusted to a range of $[0, 100]$, keeping $-1$ as an indicator of missing values. Both signals were subsequently normalised to a range of $[0.0, 1.0]$ to ensure a uniform input scale for the neural network. CTG signals of variable lengths were standardised to a fixed one-hour duration consisting of $L = 960$ time steps. Longer CTG traces were segmented into 60-minute windows, while shorter traces were padded to ensure a consistent input length. This standardisation facilitated efficient model training and ensured all input data had the same temporal length. 

The final preprocessed dataset used to train and validate PatchCTG consisted of 20,589 CTGs, comprising 10,890 NPO traces (controls) and 9,699 APO traces (cases). The 80-20 split resulted in 16,471 traces for training and 4,118 traces for validation and testing. The balanced cohorts achieved through propensity score matching ensured that the training and validation sets were free from significant biases, with Standardised Mean Differences (SMD) less than 0.10 for all controlled factors. This cohort balancing step is crucial for enabling robust performance evaluation and minimising the risk of confounding factors during model training and testing.

\subsection{Hyperparameter Optimisation}

The performance of deep learning model heavily depends on the appropriate selection of hyperparameters. Therefore, a comprehensive hyperparameter optimisation process was conducted to determine the best configuration for the PatchCTG model to accurately classify CTG signals into binary outcomes (adverse or normal) in terms of Area Under the Curve (AUC). The hyperparameter optimisation aimed to enhance model generalizability while mitigating overfitting, with the objective of maximizing the validation AUC metric. We employed the Optuna hyperparameter optimisation framework~\cite{akiba2019optuna} to perform an efficient and systematic search across a wide hyperparameter space. The optimisation process was formulated as a Bayesian optimisation problem, allowing us to iteratively explore the search space and focus on promising hyperparameter combinations based on previous trials. The goal was to identify the best hyperparameters that achieve the highest AUC score on the validation set, ensuring reliable binary classification of adverse pregnancy outcomes. The hyperparameter search covered various components of the PatchCTG architecture, including:
\begin{itemize}
    \item Transformer Encoder Layers: The number of encoder layers was varied from 3 to 6 to determine the optimal model depth that effectively captures temporal dependencies without leading to overfitting.
    \item Attention Heads: The number of attention heads was tuned from the set $\{4, 8, 16, 32\}$ to evaluate the impact of multi-head attention mechanisms on capturing complex temporal relationships.
    \item Model Dimension: The embedding dimensionality was tuned from the set $\{64, 128, 192, 256, 384, 512, 640\}$, where higher dimensionality allowed for richer feature representations, while lower dimensionality reduced computational cost.
    \item Feedforward Layer Dimension: The hidden layer dimensionality within the feedforward network was adjusted from the set $\{128, 192, 256, 320, 384, 512, 640\}$ to balance the expressiveness and complexity of the model.
    \item Dropout Rates: Dropout rates for different components (transformer layers, fully connected layers, and attention heads) were tuned in the range of $[0.1, 0.5]$ to control overfitting and improve model robustness.
    \item Learning Rate: The learning rate was selected from the set $\{1\text{e-6}, 5\text{e-6}, 1\text{e-5}, 5\text{e-5}, 1\text{e-4}, 5\text{e-4}, 1\text{e-3}\}$ to identify the most suitable rate for efficient convergence of the model.
    \item Batch Size: The batch size was varied from $\{16, 32, 48, 64\}$ to determine the optimal trade-off between convergence stability and computational efficiency.
    \item Patching Parameters: The patch length and stride for sequence patching were tuned from the set $\{4, 8, 16, 32\}$ and $\{4, 8, 16\}$, respectively, to explore different levels of sequence compression and overlapping temporal regions.
    \item Activation Function: Activation functions $\{\text{ReLU}, \text{GELU}, \text{ELU}\}$ were evaluated to determine which non-linearity yielded the most expressive feature representations for the CTG data.
\end{itemize}

We conducted hyperparameter tuning over 100 trials using Optuna, with each trial representing a unique combination of hyperparameters. Each trial was configured to train the model for a maximum of 60 epochs, with early stopping employed to halt training if no improvement was observed in validation AUC for 10 consecutive epochs. This approach ensured computational efficiency while avoiding overfitting. The validation set, consisting of 20\% of the preprocessed dataset, was used to evaluate model performance during each trial, with the AUC score serving as the primary evaluation metric. After performing the hyperparameter optimisation, the following hyperparameters yielded the highest validation AUC of 77\%:

\begin{itemize}
    \item Number of Encoder Layers: 6
    \item Number of Attention Heads: 4
    \item Model Dimension: 512
    \item Feedforward Layer Dimension: 128
    \item Dropout Rate: 0.1
    \item Fully Connected Layer Dropout: 0.4
    \item Attention Head Dropout: 0.2
    \item Patch Length: 16
    \item Stride: 16
    \item Kernel Size: 15
    \item Activation Function: ReLU
    \item Batch Size: 48
    \item Learning Rate: $1 \times 10^{-4}$
\end{itemize}

The identified hyperparameter set highlights the importance of model depth, patch length, and dropout rates in achieving high performance. Specifically, the use of six encoder layers, a modest number of attention heads, and regularisation through dropout were key factors contributing to the effective learning of temporal dependencies without overfitting, thereby improving the model's ability to generalise across different CTG signals.

\subsection{Training, Finetuning and Testing}

The PatchCTG model underwent a comprehensive training, finetuning and evaluation procedure to assess its effectiveness in classifying CTG signals into APO and NPO classes. The process involved training on a large, balanced dataset, finetuning on specific subsets and assessing generalizability and performance stability across different configurations and cohorts. The performance of PatchCTG was primarily evaluated using the Area Under the Curve (AUC) metric, along with other classification metrics such as sensitivity, specificity, Positive Predictive Value (PPV), Negative Predictive Value (NPV), F1 score and accuracy.

The initial training phase involved the entire dataset, consisting of 20,589 CTGs (10,890 NPOs and 9,699 APOs), using the hyperparameters that were optimised through the Optuna framework. The PatchCTG model was trained for a total of 50 epochs, with an early stopping criterion based on validation AUC, to ensure that the model generalised well without overfitting. The training-validation split was performed with an 80-20 ratio, providing 16,471 samples for training and 4,118 for validation and testing.

\begin{figure}[h]
\centering
\includegraphics[width=0.8\textwidth]{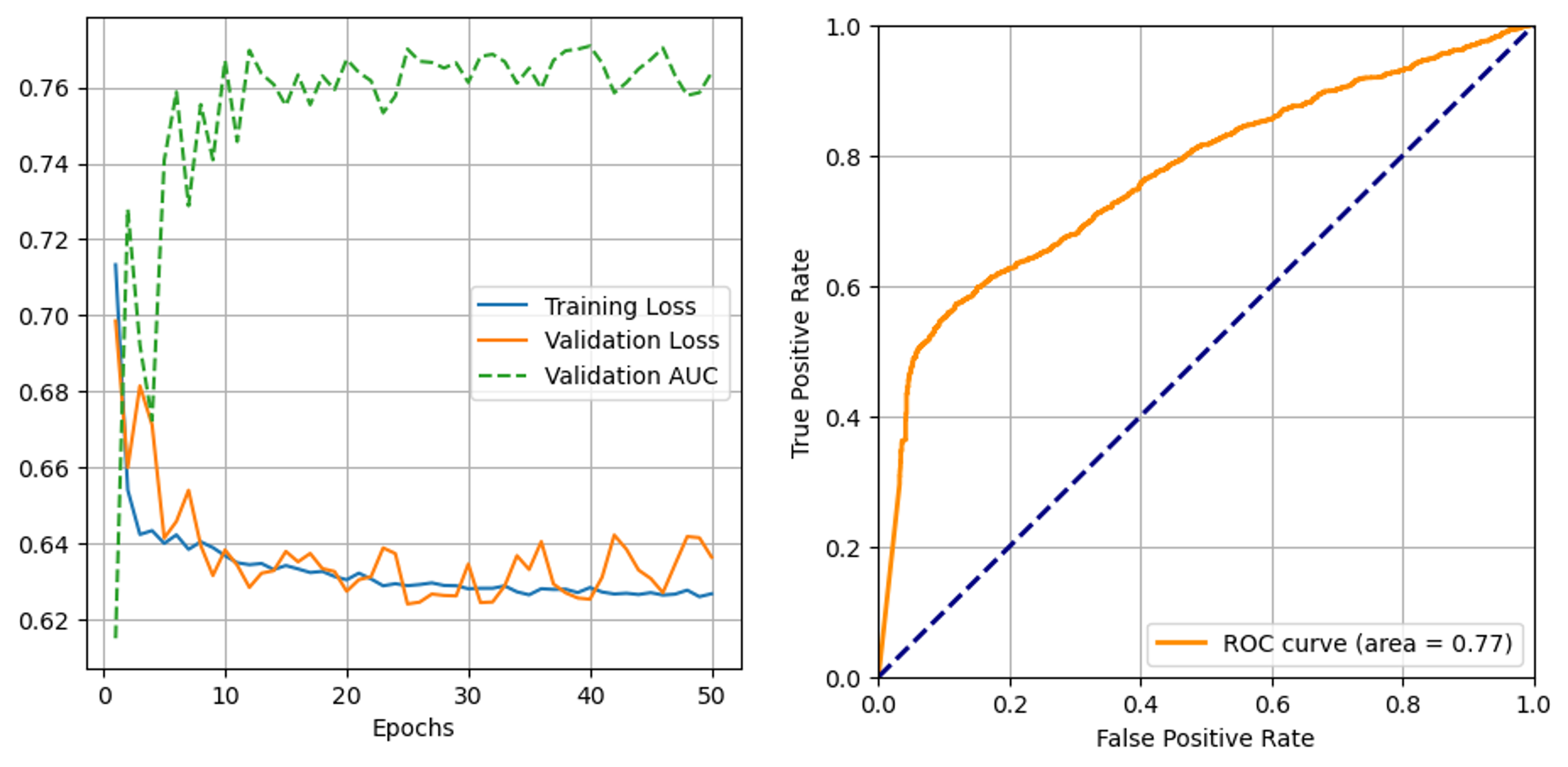}
\caption{Training and validation results of the PatchCTG model on the complete dataset with optimised hyperparameters, showing convergence to approximately 77\% AUC and consistent model performance throughout training.}\label{fig1}
\end{figure}

Figure~\ref{fig1} presents the training and validation progress plots, showing the convergence to an AUC of approximately 77\%. The ROC curve (Figure~\ref{fig1}) demonstrates the ability of PatchCTG to distinguish between adverse and normal outcomes, with the AUC well above the random guess. The performance metrics obtained for different classification thresholds, including default threshold, Youden's index threshold, high sensitivity threshold and high specificity threshold, are presented in Figure~\ref{fig3}. These results indicate a well-balanced trade-off between sensitivity and specificity, depending on the threshold selected. At Youden's index threshold, PatchCTG achieved a sensitivity of 57\%, specificity of 88\%, PPV of 81\%, and an F1 score of 67\%, which highlights the robustness of the model for clinical decision support.

\begin{figure}[h]
\centering
\includegraphics[width=0.8\textwidth]{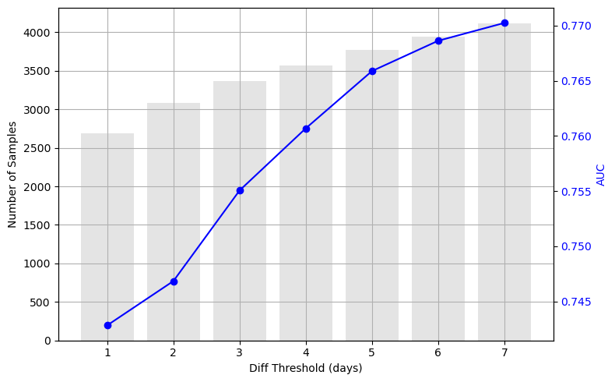}
\caption{The performance of PatchCTG in terms of AUC with varying thresholds of days to delivery (1-7) for the APO cohort, indicating a gradual improvement in AUC as the temporal threshold expands.}\label{fig2}
\end{figure}

\begin{figure}[h]
\centering
\includegraphics[width=0.8\textwidth]{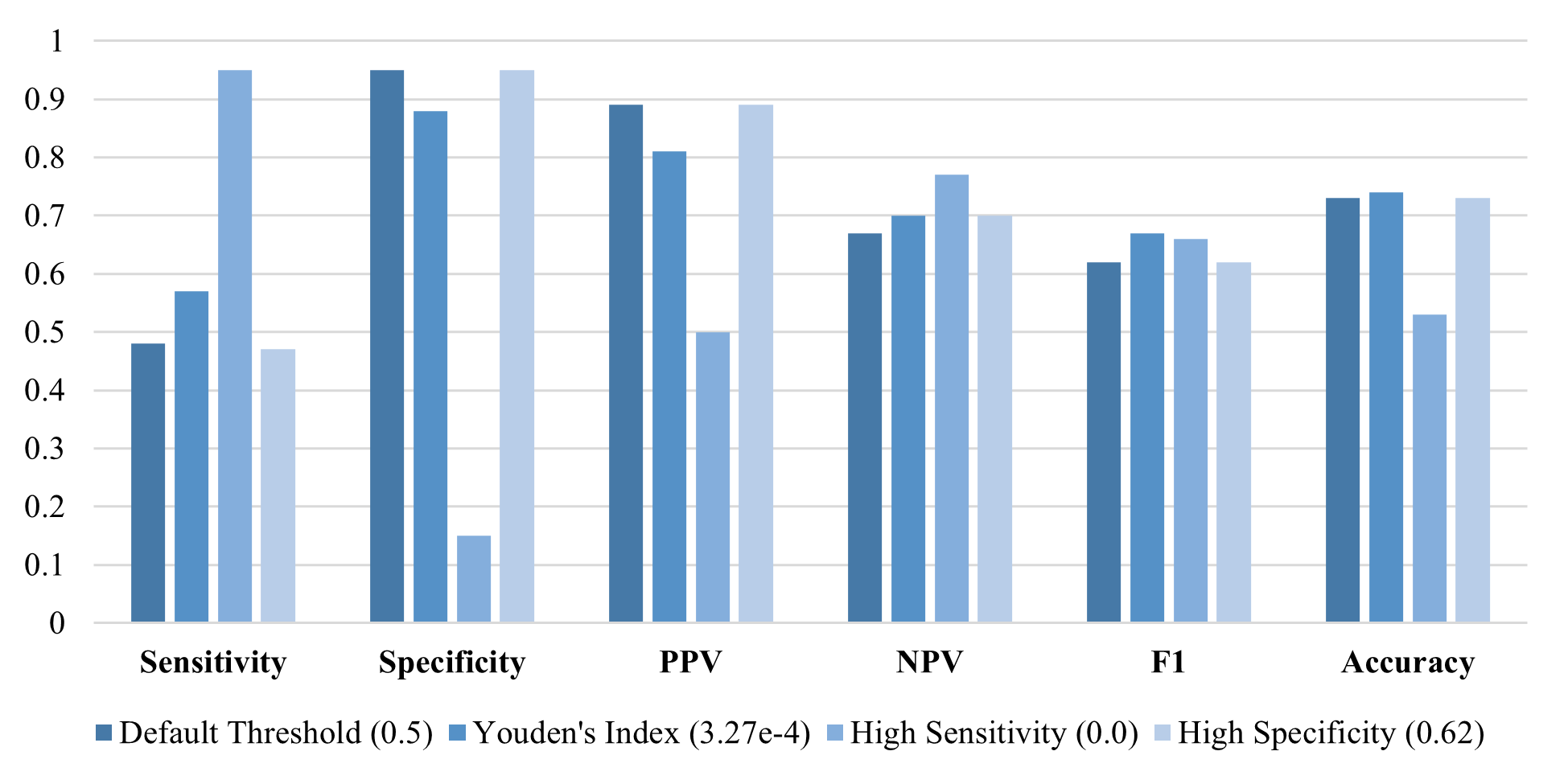}
\caption{Performance of PatchCTG model trained and tested on the complete dataset with cases recorded up to 7 days prior to delivery, evaluated using sensitivity, specificity, PPV, NPV, F1 score, and accuracy for different classification thresholds.}\label{fig3}
\end{figure}

To evaluate the impact of the temporal gap between CTG recording and delivery outcome, the PatchCTG model was assessed with varying thresholds of days to delivery for the APO cohort. Specifically, the model was evaluated on subsets of the dataset with APO cases recorded within 1 to 7 days before delivery (Figure~\ref{fig2}). The AUC increased from approximately 74\% to 77\% as the threshold increased from 1 to 7 days before delivery. The results indicated that CTG signals collected closer to delivery had a slightly lower predictive power compared to those recorded over a longer duration preceding delivery. This could be due to increased variability and abrupt changes in physiological patterns closer to delivery, which might be more challenging for the model to predict accurately.

\begin{figure}[h]
\centering
\includegraphics[width=0.8\textwidth]{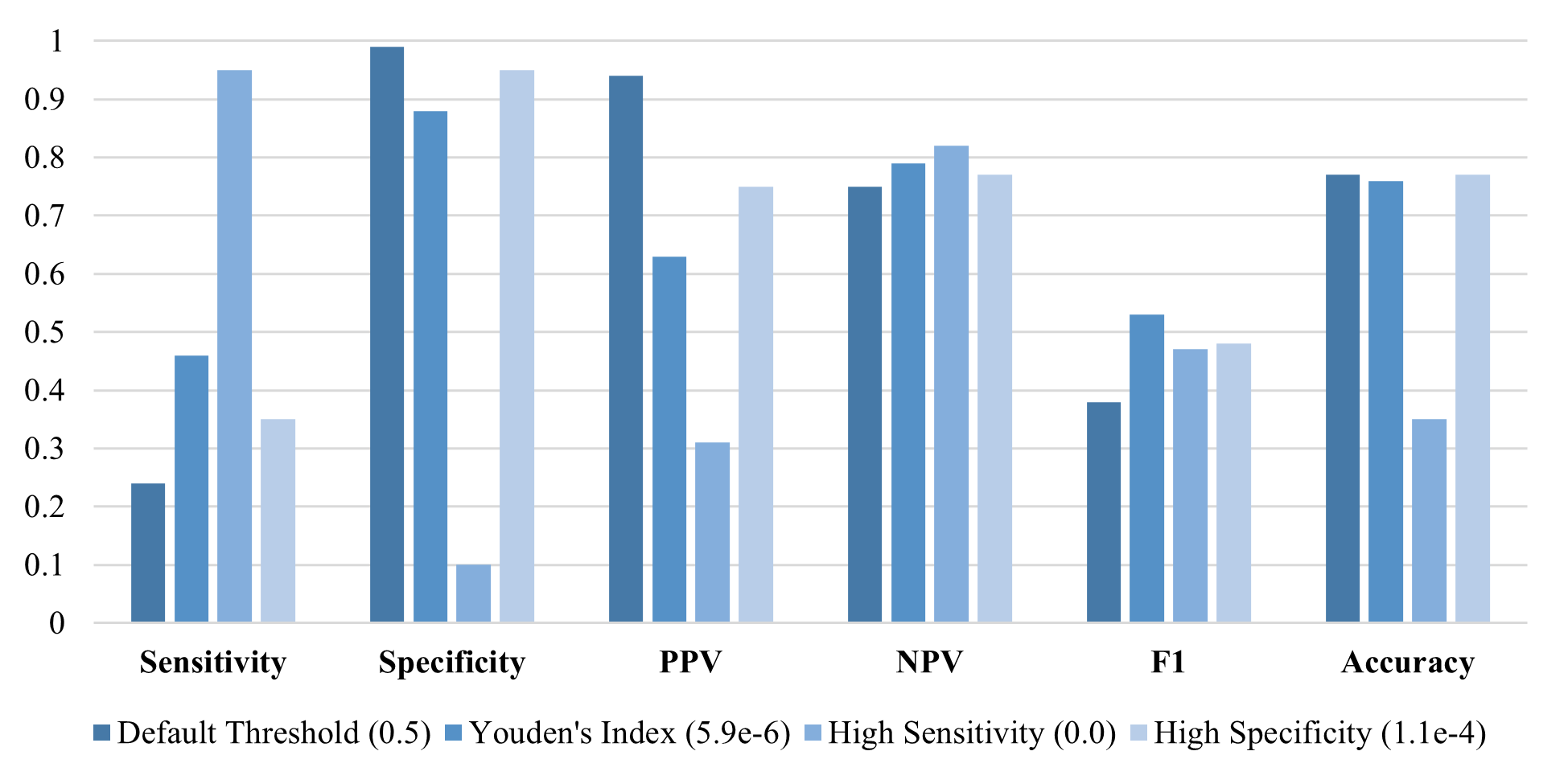}
\caption{Results of PatchCTG model trained on a subset of data with cases recorded 3-7 days prior to delivery and tested on a subset with cases recorded up to 2 days prior to delivery (AUC = 72.5\%).}\label{fig4}
\end{figure}

\begin{figure}[h]
\centering
\includegraphics[width=0.8\textwidth]{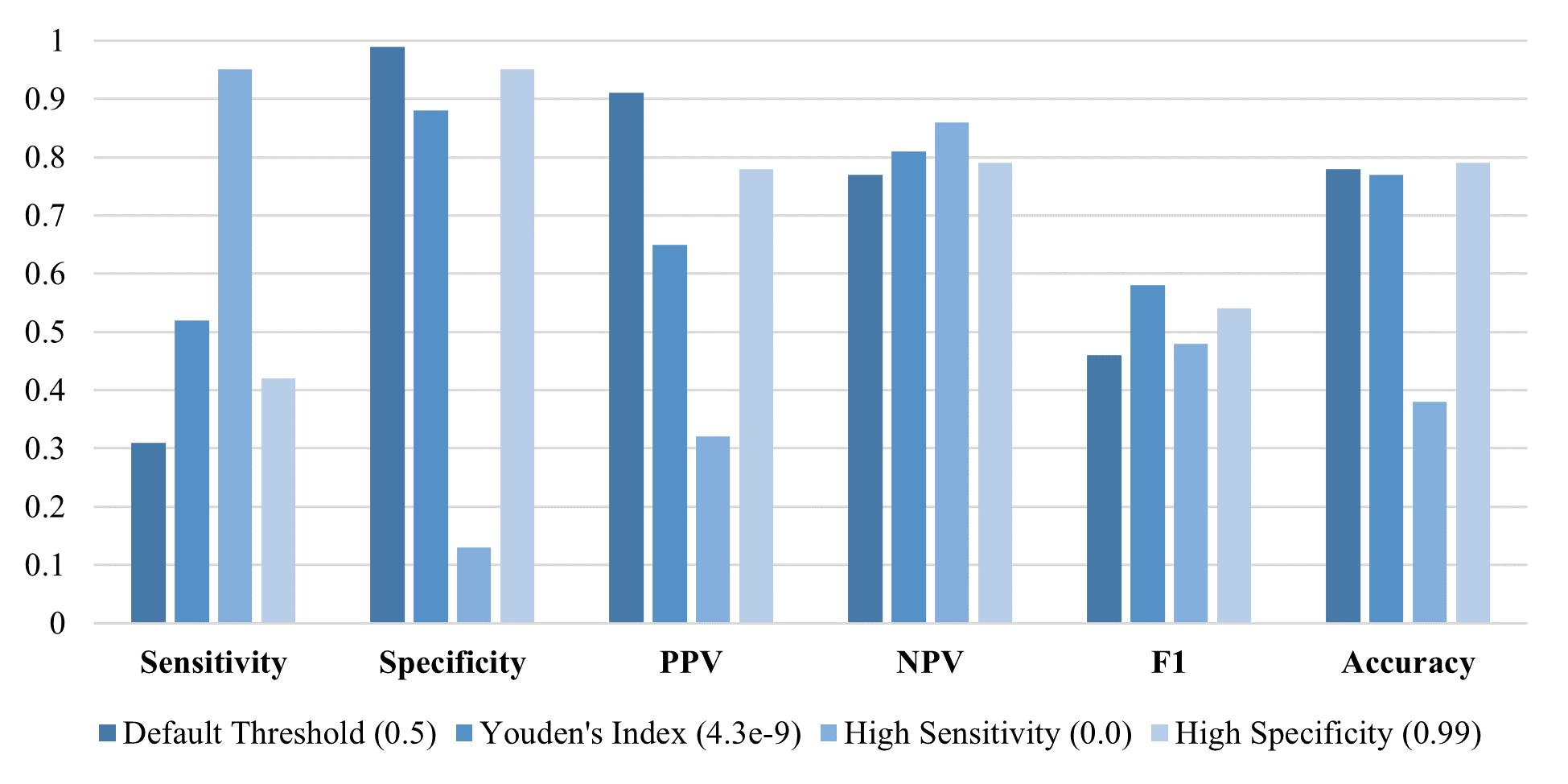}
\caption{Results of PatchCTG model pretrained on a subset of cases recorded 3-7 days before delivery, followed by finetuning and testing on cases recorded up to 2 days prior to delivery (AUC = 74.6\%).}\label{fig5}
\end{figure}

To further validate the generalizability of PatchCTG, we employed a pretraining and finetuning strategy, aimed at adapting the model to new temporal subsets of the data. The model was first pretrained using CTG signals from cases recorded 3 to 7 days before delivery and then finetuned and evaluated on a subset with cases recorded within 2 days before delivery (Figures~\ref{fig4} and~\ref{fig5}). This approach aimed to assess how pretraining on a temporally broader subset could enhance prediction performance on cases closer to delivery. During the pretraining phase, the PatchCTG model achieved an AUC of 72.5\% when trained solely on the subset of cases recorded 3 to 7 days before delivery. The metrics at various thresholds indicated that while specificity remained high at 99\% for the default threshold, sensitivity was relatively lower at 24\%, reflecting the need for further adaptation to predict outcomes more accurately when applied to a different temporal window.

Following pretraining, the model was finetuned using CTG signals from APO cohort recorded within 2 days before delivery, with results presented in Figure~\ref{fig5}. This finetuning resulted in a performance boost, with the AUC improving to 74.6\%. This showed that adapting the model to the specific temporal characteristics of the test cohort improved its predictive accuracy. Specifically, sensitivity increased from 31\% to 52\% at the Youden's index threshold, while specificity remained high at 88\%, showing that the model successfully adapted to temporal shifts in the data.

\begin{figure}[h]
\centering
\includegraphics[width=0.8\textwidth]{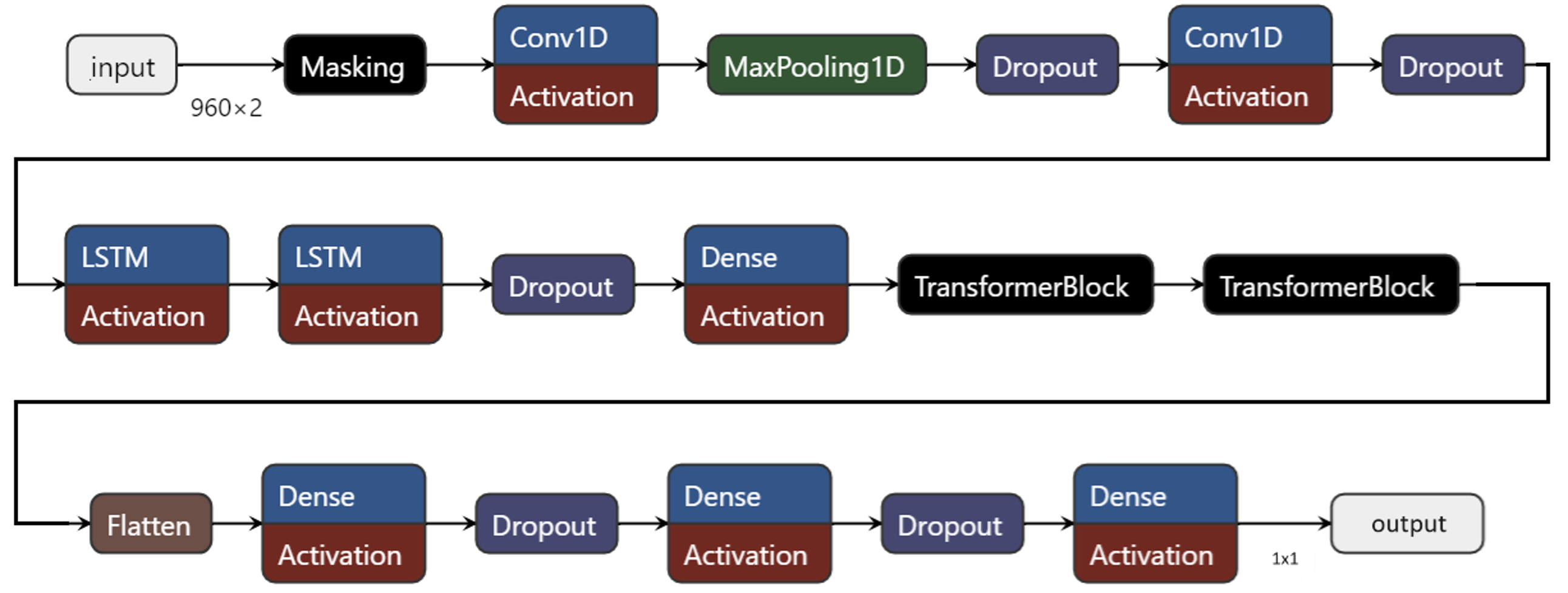}
\caption{Hybrid deep learning model identified through hyperparameter tuning for benchmark comparison with PatchCTG}\label{fig6}
\end{figure}

\subsection{Benchmark Evaluation}
The benchmark comparison evaluates the performance of PatchCTG relative to the DR algorithms and a hybrid deep learning model for classifying antepartum CTG data into adverse and normal outcomes. The hybrid deep learning model was optimised via an extensive hyperparameter tuning process, which explored variations in CNN, LSTM and Transformer layers to identify the most effective architecture for CTG classification. Hyperparameters for CNN layers included up to three layers with varying filters, kernel sizes, L2 regularisation, activation functions, and dropout rates. The LSTM layers were tuned for unit count, regularisation, and dropout, while Transformer blocks were tested with different embedding dimensions, feedforward dimensions, attention heads, and block counts. The optimal architecture (shown in Figure \ref{fig6}) identified through this process included two CNN layers, three LSTM layers and two Transformer blocks, achieving an AUC of 73.5\%. This model achieved a sensitivity of 72\% and specificity of 64\% at the Youden’s index threshold. In comparison, the traditional Dawes-Redman (DR) algorithm attained an AUC of 67\%, reflecting lower discriminatory power than both the CNN-LSTM-Transformer model and PatchCTG. The DR algorithm demonstrated a high specificity (90.7\%) but low sensitivity (18.2\%), correctly identifying the majority of normal outcomes but showing limited capacity to detect adverse outcomes. In contrast, PatchCTG achieved the highest AUC of 77\%, outperforming both the CNN-LSTM-Transformer model and the DR algorithm. At Youden's index threshold, PatchCTG attained a balanced sensitivity of 57\% and specificity of 88\%. PatchCTG's high AUC demonstrates its effective handling of both local and global temporal dependencies in CTG signals, attributed to its patch-based segmentation and self-attention mechanisms, which enable it to adapt across diverse temporal patterns and signal variabilities inherent in CTG data. 

\section{Discussion}\label{sec5}

PatchCTG demonstrated robust performance with an AUC of 77\%, highlighting its capacity to accurately classify CTG recordings as adverse or normal outcomes in the antepartum setting. Compared to traditional CTG analysis models, such as the Dawes-Redman system \cite{pardey2002computer} and machine learning approaches \cite{krupa2011antepartum, fei2020automatic}, PatchCTG provides a more consistent, objective evaluation, addressing long-standing issues in CTG interpretation, including inter-observer variability and limited predictive capability for adverse outcomes. This consistency is critical given the limitations of prior models, which have shown lower specificity and varying sensitivities, often requiring complex feature engineering to capture nuanced temporal patterns in FHR signals \cite{chen2021intelligent, georgieva2013artificial}. The integration of patch-based segmentation and self-attention mechanisms in PatchCTG represents a significant advancement in CTG analysis, drawing on transformer architectures that have shown promise in other medical time-series applications \cite{barnova2024artificial}. By segmenting CTG signals into patches and applying instance normalisation and channel-independent processing, PatchCTG efficiently captures both local and global temporal dependencies, which are essential for interpreting the physiological dynamics in FHR and uterine contraction signals. Unlike convolutional models \cite{petrozziello2019multimodal, fei2022intelligent} that excel in extracting spatial features but can struggle with long-range dependencies, PatchCTG leverages the self-attention mechanism to dynamically adjust its focus across signal patches, enhancing its ability to detect subtle patterns associated with adverse outcomes.

Evaluating PatchCTG under different temporal thresholds demonstrated its generalisability across various intervals before delivery, with some reduction in performance for signals recorded closer to delivery. This degradation may reflect the increased variability and subtle changes in physiological patterns as delivery approaches, highlighting the need for additional clinical markers or features to improve prediction accuracy during these critical hours. Importantly, the ability of the model to leverage broader temporal data during pretraining, with finetuning on closer-to-delivery signals, illustrates an approach beneficial for clinical settings where data from different time windows may vary in availability and relevance. The performance of PatchCTG across different classification thresholds also underscores its adaptability for clinical priorities. By adjusting the threshold to increase sensitivity, the model can be tuned to minimise false negatives, which is essential in high-risk clinical scenarios where missing an adverse outcome could lead to severe consequences. Conversely, a high specificity threshold could help reduce unnecessary interventions when the priority is to avoid false positives. This flexibility makes PatchCTG a valuable tool for aiding clinical decision-making in fetal monitoring.

Benchmark comparisons further underscore the strong performance of PatchCTG relative to an optimized hybrid deep learning model and the DR algorithm. With an AUC of 77\%, PatchCTG outperformed the CNN-LSTM-Transformer model, which achieved an AUC of 73.5\%, and the Dawes-Redman algorithm, which had an AUC of 67\%. This comparison underscores the enhanced capability of PatchCTG to capture critical temporal dependencies while maintaining high predictive accuracy, particularly compared to conventional methods that exhibit lower specificity and sensitivity trade-offs. Overall, PatchCTG addresses gaps identified in prior deep learning methods for CTG analysis by efficiently capturing temporal dependencies, reducing subjectivity and enabling adaptable outputs. Future work should explore integrating multimodal biomedical and clinical data to further enhance predictive power, particularly as delivery approaches and validation across larger, more diverse datasets to ensure model generalisability and real-world impact.

\section{Conclusion}\label{sec6}
This study introduces PatchCTG, a transformer model explicitly designed for antepartum CTG-based fetal health monitoring. Achieving an AUC of 77\%, PatchCTG outperformed the Dawes-Redman system (AUC of 67\%) and an optimised hybrid deep learning model (AUC of 73.5\%). The ability of PatchCTG to capture complex local and global temporal dependencies, along with its adaptability across varying timeframes, positions it as a valuable tool for clinical application, offering greater reliability and objectivity than traditional methods. Its adaptable sensitivity and specificity thresholds further enhance its clinical utility, allowing for precision adjustments that prioritise sensitivity in high-risk cases or specificity to minimise unnecessary interventions. This flexibility, combined with the capacity of PatchCTG to generalise across different temporal windows, supports a robust approach to CTG interpretation that can help reduce the subjectivity common in manual assessments. While the performance of PatchCTG is promising, further enhancements could be achieved by incorporating additional data sources and clinical markers, particularly to improve predictive accuracy closer to delivery. Future work will focus on expanding clinical validation across diverse datasets and exploring the integration of multimodal inputs to enhance fetal health assessment and support more timely, informed clinical decisions.

\section*{Data and Code Availability}
The source code is available at \href{https://github.com/jaleedkhan/PatchCTG}{https://github.com/jaleedkhan/PatchCTG}. The OXMAT dataset used in this work cannot be publicly shared due to ethical and safety concerns. It contains healthcare information that could reveal sensitive details, potentially compromising patient anonymity. We are committed to upholding the highest standards of privacy and confidentiality in accordance with ethical guidelines and regulations. Further details about the OXMAT dataset can be found in the paper (\href{https://arxiv.org/abs/2404.08024}{https://arxiv.org/abs/2404.08024}) and the documentation website (\href{https://oxmat.oxdhl.com/}{https://oxmat.oxdhl.com/}). Researchers, healthcare professionals and institutions interested in exploring new research questions, validating findings, refining data processing and analysis techniques or contributing to the further development of the OXMAT dataset are encouraged to contact the authors at oxmat@wrh.ox.ac.uk. 

\section*{Acknowledgements}
This work was supported by the Medical Research Council [MR/X029689/1] and The Alan Turing Institute's Enrichment Scheme. This work is indebted to the work of Profesor Christopher Redman on developing the Dawes-Redman criteria and advancing CTGs, to Beth Albert for her clinical guidance, and to Pawe\l{} Szafranski, James Bland, Ioana Duta and John Tolladay for their work compiling, anonymising and documenting the data. 

\section*{Author Contributions}
M. J. Khan was responsible for the methodology design and implementation, experimental analysis and manuscript writing. M. Vatish and G. Davis Jones reviewed the manuscript. G. Davis Jones also conducted data extraction, supervised the work and contributed to the analysis. 

\section*{Competing Interests}
The author(s) declare no competing interests.

\bibliography{sn-bibliography}

\end{document}